\documentclass{article}

\usepackage{arxiv}
\usepackage{times}
\usepackage{latexsym}
\usepackage{amsfonts}
\usepackage{amsmath}
\usepackage{multirow}
\usepackage{enumitem}
\usepackage{graphicx}
\usepackage{subcaption}
\usepackage{color}
\usepackage{booktabs}
\usepackage{bm}
\usepackage{array}
\usepackage{booktabs}
\usepackage{cprotect}
\usepackage{url}
\usepackage[utf8]{inputenc} % allow utf-8 input
\usepackage[T1]{fontenc}    % use 8-bit T1 fonts
\usepackage{hyperref}       % hyperlinks
\usepackage{booktabs}       % professional-quality tables
\usepackage{nicefrac}       % compact symbols for 1/2, etc.
\usepackage{microtype}      % microtypography

\DeclareMathOperator*{\argmin}{arg\,min}
\definecolor{yamabuki}{RGB}{248, 181, 0}

\title{Training with Streaming Annotation}
\author{
  Tongtao Zhang\\
  Siemens Corporate Technology \\
  Princeton, NJ 08540\\
  \texttt{tongtao\_D\_zhang\_A\_siemens\_D\_com} \\
   \And
  Heng Ji\\
  University of Illinois Urbana-Champaign\\
  Urbana, IL 61801 \\
  \texttt{hengji\_A\_illinois\_D\_edu  } \\
  \And
  Shih-Fu Chang\\
  Columbia University \\
  New York, NY 10025 \\
  \texttt{shih.fu.chang\_A\_columbia\_D\_edu} \\
  \And
  Marjorie Freedman\\
  Information Sciences Institute\\
  Waltham, MA 02451\\
  \texttt{mrf\_A\_isi\_D\_edu} \\
}

\begin{document}
\maketitle
\begin{abstract}
In this paper, we address a practical scenario where training data is released in a sequence of small-scale batches and annotation in earlier phases has lower quality than the later counterparts. To tackle the situation, we utilize a pre-trained transformer network to preserve and integrate the most salient document information from the earlier batches while focusing on the annotation (presumably with higher quality) from the current batch. Using event extraction as a case study, we demonstrate in the experiments that our proposed framework can perform better than conventional approaches (the improvement ranges from 3.6 to 14.9\% absolute F-score gain), especially when there is more noise in the early annotation; and our approach spares 19.1\% time with regard to the best conventional method.\footnote{We will make all resources and programs publicly available for research purpose.}

 \end{abstract}
\section{Introduction}
\label{sec:introduction}

%\heng{it's still not very convincing why this kind of training scenario (training data comes in stream) is common and useful; need to make the motivation stronger}
%\heng{overall the paper is very terse, not very engaging. you need to polish it to make it more juicy}

Successful supervised models depend on high quality annotation. Although there are platforms such as Amazon Mechanical Turk or tools such as LightTag or brat~\cite{brat2012} which facilitates ``crowd-sourcing'' annotation and help the researcher and developers to acquire large scale annotation within a short period, we do not neglect merits of hiring a professional annotation group if the data set contains credential or privacy information and/or the task requires intensive training on background knowledge of expertise domains.

In an ideal setting, professional annotation construction is performed with the following steps:
\begin{enumerate}[noitemsep,nolistsep,leftmargin=*]
\item The task (e.g. set of a labels) is defined; 
\item Task relevant data is collected;
\item Annotators are recruited and trained;
\item Annotators annotate the data, and adjudicate the annotation. 
\end{enumerate}

In practice, the above procedure is repeatedly iterated, \textit{i.e.,} data sponsors or sources provide additional data, annotators review the previous data and annotation, and adjudicators correct errors or resolve disagreements. This process is usually lengthy, e.g., ACE2005 has $600$ documents and $15,000$ sentences and it took more than three years for them to be fully annotated and adjudicated~\cite{walker2006ace}. From system developers and researchers' perspective, waiting for a large corpus to be fully annotated is impractical.

%%From Marjorie: Heng had asked for better motivation, some thoughts on this:
%%% 1) At least in current results section, it seems like biggest strength is grappling with data that starts out noisy and gets less noisy over time. It might be stronger to make this claim more assertively at the beginning. 
%%% 2) you could consider a motivating use case of "the task designers get more information (e.g. discover that the the initial definitions were confusing, the situation evolves and they adjust label guidelines to support that change.  
%%% 3) This doesn't fit in with the notion of noise being reduced over time, but a second good motivating use-case is domain drift. If you have an important problem (and are in production) you might want to be constantly updating your annotations so the the model does not suffer from out-of-vocabulary content.  e.g. the ACE data is old enough that Facebook and Twitter are unknown words. If accurate NER were in production, you could imagine wanting to be constantly adding to the data set to support new entities. If you wanted to extend the paper, the genres of the ACE data (BC, BN, NW, etc. could be used a proxy for the deomain shift.

To shorten the waiting time, annotation organizers release annotation in small batches -- \textit{\textbf{stream}} -- so that system developers and researchers are able to get familiar with the early release and start to work. This requires domain experts to carefully define and confirm the schema. Moreover, with limited budget, time constraint and work load balance, it is very likely that annotations among batches yield to different quality. Usually the quality in early batches is lower than that in later counterparts due to various reasons, \textit{e.g.,} annotators and adjudicators may misunderstand the schema, or they may be unfamiliar with the annotation tools and make mistakes in early stages, and they may not have sufficient efforts or time to re-visit and re-annotate those early data sets while focusing on later batches with more skillful annotation.

\begin{figure*}[ht]
\centering
\includegraphics[width=\textwidth]{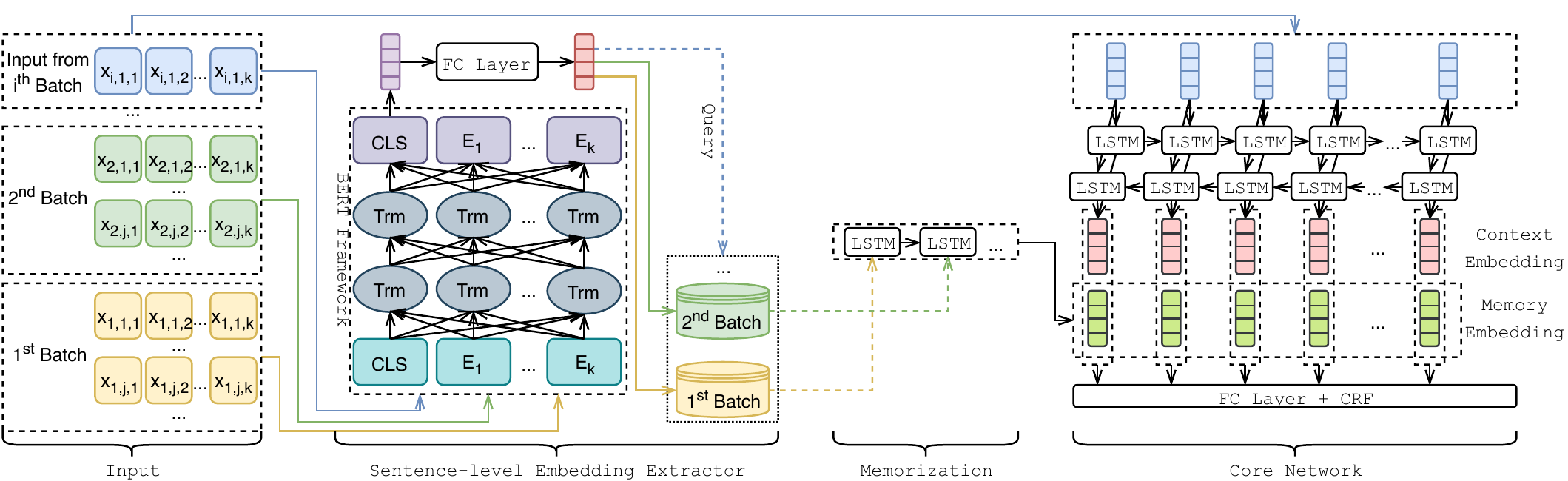}
\caption{An overview of the proposed framework. For an input sentence in the $i^{th}$ batch, it retrieves the most similar sentence-level embeddings from the previous batches -- one embedding from one batch (Section~\ref{sec:bert}). The sentence-level embeddings from previous batches go through a forward LSTM network which outputs memory embeddings (Section~\ref{sec:memory}). Memory embeddings are then concatenated with the context embedding from the Bi-LSTM and fed into the CRF model. (Section~\ref{sec:core}).}
\label{fig:memory_framework}
\end{figure*}

Data and annotation available in stream-style might be handled by the following strategies:
\begin{enumerate}[noitemsep,nolistsep,leftmargin=*]
\item When a new batch of training data arrives, aggregate them with the old batch and train a new model.
\item Train a model with the first batch, and when a new batch arrives, finetune the model with the new batch.
\item When a new batch of training data arrives, train a new model with that batch and dump the models from previous batches.
\end{enumerate}
The first strategy assumes that the data and annotation have already been delivered as a complete corpus and consumes all the data available, this strategy will inevitably introduce errors from the data in the earlier batch. The second strategy is also risky since the distribution of errors is unknown and newly arrived data does not ensure an efficient suppression on those error annotations. Although the third strategy seems promising because the later batches will come with better annotation, it usually brings an over fitted model especially when size of batches is small. Hence, we require a paradigm that handles streaming scenario by efficiently capturing information from annotation in current batch and similar sentences in early batches and alleviate errors in early annotation.

%Some approaches assume that the data and annotation have already been delivered as a complete corpus and consume all available data and train them iteratively, and on each 

%The original approach to handle streaming data is accumulating new batches and training new models from aggregated batches. However, such method will preserve the distribution from old data and impact the current batch; moreover, training time is inevitably prolonged and old models are wasted.

According to our empirical practises with and as annotators, we notice that the improvement in the annotation quality with the progress going on depends on a few reasons. Besides a better understanding in the annotation schema and skilling up in operating the annotation tool, annotators' accumulated impression on the passed documents also helps with improved annotations. For example, when seeing the word ``\textit{strike}'', which may be a trigger word of \verb|Attack| event or a trigger word of \verb|Demonstrate| event, skillful annotators immediately determine an ``\verb|Attack|'' label on it if the target sentence resembles those passed documents with country and military forces names, or a ``\verb|Demonstrate|'' label if the target sentence resembles those ones with business names and individual employee names.

In this paper, to emulate the aforementioned process, we propose a framework that capture the sentence-level features extracted from old stream batches, whose annotations are ignored by the core networks and the core network focuses on the annotation from the current batch. This framework will ensure one single model which is continuously finetuned mainly by the current batch while preserving the most informative features from old data and suppressing errors in the past batches. We use event extraction as a study case. The event annotation and the extraction framework follow \textit{ACE}\footnote{\textbf{A}utomatic \textbf{C}ontent \textbf{E}xtraction, \url{https://catalog.ldc.upenn.edu/LDC2006T06}} schema where 33 event types (such as \verb|Attack| and \verb|Meet|) are defined. The extraction system aims to extract \textit{event triggers}, which mostly express events (\textit{e.g.}, ``\textit{shoot}'' for \verb|Attack|). We assume the following conditions in the study:
\begin{enumerate}[noitemsep,nolistsep,leftmargin=*]
%\item Schemas are fixed, and there are no more changes in the definition in the labels.
\item Once annotation and adjudication are done on a data instance, annotators and adjudicators \textbf{do not} \textit{re-visit} the instance.
\item The quality of annotation is lower in early batches and increases in the later ones.
\end{enumerate}

The contributions of the paper are summarized as follows:
\begin{enumerate}[noitemsep,nolistsep,leftmargin=*]
\item We propose a new approach that tackles annotated data that arrives in small batches, and the proposed approach is robust against annotation errors among the batches.
\item To the best of our knowledge, the proposed framework is the first to explore a scenario where annotation data arrives in small-scale batches, especially for event extraction task. 
\end{enumerate}

% \item When a new batch of training data arrives, aggregate them with the old batch and train a new model. In this strategy, all models include all annotation errors.
% \item Train a model with the first batch, and when a new batch arrives, finetune the model with the new batch. With this strategy, it seems that the influence of the early errors are alleviated, however, biased and volatile distribution in the data still affects the performance.
% \item When a new batch of training data arrives, train a new model with that batch. Although this strategy avoids errors, a model trained with a small batch easily overfits.
\section{Approach}
% Component usage
% Component input and component output
% Component composition
% How do you train the component
\subsection{Sentence-level embeddings}
\label{sec:bert}
We retrieve the most similar sentences from the previous batches in terms of the \textit{existence of same event types} for an input sentence. Additional features and embeddings from these sentences improve event trigger extraction performance of the core network (Section~\ref{sec:core}).

Given a sentence $X_{i,j}$ where $i$ denotes the index of a batch and $j$ denotes the index within the batch. We have
\begin{equation}
\bm{x}_{i,j} = F_s(X_{i,j})
\end{equation}
where $F_s(\cdot)$ denotes a sentence-level embedding extractor. As shown in Figure~\ref{fig:memory_framework}, we use the BERT\cite{devlin2018bert} embedding from the ``\verb|CLS|'' token (\verb|CLS| stands for \textit{class}, this token is used for sentence classification) and we append a fully-connected layer after the embedding as a sentence-level feature extractor. We will introduce how we train this additional FC layer in the experiment section.

The sentence-level embeddings are stored based on the batch tag of the source sentence. For the $i^{th}$ batch, we denote
\begin{equation}
\tilde{\mathcal{X}}_i = \{\bm{x}_{i,1}, \ldots, \bm{x}_{i, j}, \ldots\},
\end{equation}
where $j$ denotes the index of a sentence in this batch.

For each input sentence embedding $\bm{x}_{i,j}$, we retrieve the closest embedding from each of the previous batches. We denote
\begin{equation}
\hat{\mathcal{X}}_{i,j} = \{ \bm{x}_{1, \hat{j}_1}, \bm{x}_{2, \hat{j}_2}, \ldots, \bm{x}_{m, \hat{j}_{m}}, \ldots, \bm{x}_{i-1, \hat{j}_{i-1}}\},
\end{equation}
where the footnote $\hat{j}_m$ denotes the index of the closest sentence-level embedding in the $m^{th}$ batch and $1\leq m \leq i-1$:
\begin{equation}
\hat{j}_m = \argmin_{j_m} dist(\bm{x}_{i,j}, \bm{x}_{m, j_m}).
\end{equation}
We define the distance function $dist(\cdot, \cdot)$ as the $L^2$ or Euclidean distance.

\subsection{Memory Embedding}
\label{sec:memory}
Starting from the second batch, we feed the queried sentence-level embedding $\hat{\mathcal{X}}_{i,j}$ from the previous batch(es) into a forward LSTM (Long short-term memory)~\cite{hochreiter1997long} network. Each previous batch provides one sentence-level embedding and these embeddings form a sequence input to the LSTM network. The output, or the final state of the LSTM network is a memory embedding $\mathcal{M}_{i,j}$. 

%The memory network is a forward LSTM (Long short-term memory)~\cite{hochreiter1997long} network. It accepts the queried sentence-level embeddings $\hat{\mathcal{X}}_{i,j}$ in a sequence. The final state of the LSTM network is considered as a \textit{Memorization Embedding} $\mathcal{M}_{i,j}$.

% In the $m$th \heng{make all these 'th' upper-script} batch, when a sentence arrives, the sentence \heng{do you mean the model?} goes through the sentence level feature extractor, and the vector embedding will query the most similar sentence embeddings in the earlier batches. The retrieved sentence embeddings followed by the current sentence embedding will be fed into the memory network. The output is the last hidden state from the LSTM network, and will be concatenated with context embeddings in Section~\ref{sec:core}.

This submodule simulates how human annotator recalls similar sentences that contain the same event types in the past documents when he/she is working on a current sentence. For example, we use the sentence embedding for ``\textit{Putin last spoke to Bush on April 5 at the US president's own initiative}'' as a query and retrieve the sentence ``\textit{Putin last visited Bush at his Texas ranch in November 2001}''  which contains similar entities and related event (\verb|Meet| event triggered by ``\textit{visited}''). The memory embeddings calculated from the sentence embeddings trough the LSTM network provide the most salient information to enhance the detection of target labels such as event triggers.

\subsection{Core network}
\label{sec:core}
The core network is based on a Bi-LSTM-CRF framework~\cite{ma2016end,feng2018language}, which models trigger extraction as a sequence labeling problem.

We feed the current sentence which is tokenized into a Bi-LSTM network and acquire the context embeddings consisting of the concatenated hidden output from each token step. In the ``vanilla'' version of this framework, the context embeddings are then fed into an FC-layer and CRF layer to generate the final output. In our framework, according to Figure~\ref{fig:memory_framework}, starting from the second batch where we are able to retrieve sentence and memory embeddings from previous batch(es), we concatenate the memory embedding $\mathcal{M}_{i,j}$ with each word contextual embedding before we feed them into the FC layer and the CRF (conditional random field) layer.
\section{Experiments}
\label{sec:memory_experiments}
\subsection{Data and Simulation}
\label{sec:data_split}
Since most publicly available data sets have already been carefully annotated and adjudicated and we are not able to access and recover the annotation history, to evaluate the performance with our proposed framework, we simulate the scenario of increasing quality by randomly dropping or swapping the labels. In this section, we use the ACE2005 data set, which follows ACE schema and contains 33 types of event annotation, as the test bed. In comparison with other sequence labeling data sets, it has a substantially large label space (67 considering BIO\footnote{In BIO settings, B denotes the \textit{beginning} of a word/phrase segment, I denotes \textit{inside} the segment, \textit{e.g.}, the phrase ``\textit{shot down}'' is labeled as ``\textit{shot}'' with B-Attack and ``\textit{down}'' with I-Attack. We have 2 labels for each event type and 33 event types provide 66 labels, and we also have an O label to denote a token without any event label, hence we have 67.}). We select documents from \verb|bc| (broadcast conversations), \verb|bn| (broadcast news) and \verb|nw| (newswire). These documents are tagged with release dates between March and June 2003. We simulate the batches according to the months and list the batch information in Table~\ref{tab:ace_stats}. For each batch, we split the data into training, development and test set as shown in Table~\ref{tab:split_ace2005}.

We simulate annotation errors by noise levels, which indicate the probability of swapping and dropping labels, in the training sets which decreases along with the month tags. We have two settings of simulation, and we denote the settings using the noise level of the first month, and we have noise levels for each month tag as follows:
\begin{itemize}[noitemsep,nolistsep,leftmargin=*]
\item \textbf{25\%}: 25\%, 10\%, 5\%, and 0\%.
\item \textbf{10\%}: 10\%, 5\%, 0\%, and 0\%.
\end{itemize}
In the tables, we use ``\textit{Noise}'' represent the two simulation groups.

We use spaCy~\footnote{https://spacy.io} to acquire the stem and lemmatized string of each token in the training set, and then we have a ``confusing'' list of trigger stems with multiple event type labels. For example, ``\textit{fire}'' can be an \verb|Attack| trigger or a \verb|End-Position| trigger. When simulating the erroneous annotation, we set a threshold based on the noise level for that month, and if the random generator generates a number lower than the noise level number, we either drop the label for that trigger (or we just make the label as \verb|O|), or if the trigger appears on the aforementioned ``confusing'' list, we randomly select another event label which also includes this trigger (and we can also set a \verb|O| label). These actions simulate the scenario where annotators ``miss'' (with \verb|O| label) or ``are confused'' with other event types that the word may also trigger ((with other labels). However, for the sake of valid parameter tuning and fair comparison, the labels in the development and test sets remain intact.

For each noise level, we generate 10 groups of simulation data, and we run experiments 10 times and calculate the average of precision and recall scores, then calculate F1 scores.

\begin{table}[t]
\centering
\begin{tabular}{c|cccc}
\toprule
Months & Doc. & Sent. & Words & Triggers\\
\midrule
200303 & $67$ & $2,578$ & $39,825$ & $960$ \\
200304 & $148$ & $3,834$ & $62,443$ & $1,406$ \\
200305 & $76$ & $1,566$ & $28,542$ & $641$\\
200306 & $92$ & $1,748$ & $28,315$ & $578$ \\
\bottomrule
\end{tabular}
\caption{The stat of ACE2005 with month tags.}
\label{tab:ace_stats}
\end{table}

\begin{table}[t]
\centering
\begin{tabular}{c|cc|cc}
\toprule
Months & Train & Dev & Test & All Test\\
\midrule
200303 & $53$ & $7$ & $7$ &\multirow{4}{*}{$40$}\\
200304 & $118$ & $15$ & $15$ &\\
200305 & $60$ & $8$ & $8$  &\\
200306 & $73$ & $9$ & $10$ &\\
\bottomrule
\end{tabular}
\caption{Training, development and test splits with month tags. For intuitive comparison, we use all 40 documents as the test set regardless of their month tags.}
\label{tab:split_ace2005}
\end{table}

\begin{table*}[t]
\centering
\begin{tabular}{c|c|ccc|ccc|ccc|ccc}
\toprule
  \multirow{2}{*}{Noise} & Slice &\multicolumn{3}{c|}{200303}&\multicolumn{3}{c|}{200304}&\multicolumn{3}{c|}{200305}&\multicolumn{3}{c}{200306}\\
  & Metric & P & R & F1 & P & R & F1 & P & R & F1 & P & R & F1\\
\midrule
\multirow{4}{*}{25\%} & \textit{All} & 56.9 & 56.6 & 56.7 & 63.4 & \textbf{68.7} & 65.9 & 70.1 & 68.6 & 69.3 & 71.3 & 70.4 & 70.8 \\
& \textit{Current} & 56.9 & 56.6 & 56.7 & 60.5 & 55.8 & 58.1 & 60.4 & 48.3 & 53.7 & 63.5 & 56.0 & 59.5 \\
  & \textit{Finetune} & 56.9 & 56.6 & 56.7 & 67.9 &  67.7 & 67.9 & 66.6 & 64.8 & 65.7 & 70.2 &  66.5 & 68.3 \\
  & \textit{Proposed} & 56.9 & 56.6 & 56.7 & \textbf{74.2} & 66.7 & \textbf{70.3} & \textbf{71.2} & \textbf{74.1} & \textbf{72.7} & \textbf{72.9} & \textbf{76.1} &\textbf{74.4}\\
\midrule
\multirow{4}{*}{10\%} & \textit{All} & 67.9 & 70.4 & 69.1 & 80.1 & 78.3 & 79.2 & 82.7 & \textbf{79.9} &81.2 & 83.4 & 82.3 & 82.8\\
& \textit{Current} & 67.9 & 70.4 & 69.1 & 64.8 & 60.8 & 62.7 & 54.4 & 57.2 & 55.8 & 63.5 & 56.0 & 59.5\\
& \textit{Finetune} & 67.9 & 70.4 & 69.1 & 72.7 & 75.6 & 74.1 & 77.3 & 74.8 & 76.0 & 74.5 & 76.7 & 75.5\\
& \textit{Proposed} & 67.9 & 70.4 & 69.1  & \textbf{80.5} & \textbf{79.6} & \textbf{80.1} & \textbf{85.7} & 78.9 & \textbf{82.2} & \textbf{84.5} & \textbf{82.8} & \textbf{83.6}\\
\bottomrule
\end{tabular}
\caption{Performance (\%) comparison with different strategies and noise level. Names of strategies (\textit{All}, \textit{Current}, \textit{Finetune}, \textit{Proposed}) and definition of noise level are introduced in Section~\ref{sec:memory_settings}. Note that there is no ``previous'' batch for the first batch, the results in the first batch are identical.}
\label{tab:performance_ace2005}
\end{table*}

\subsection{Settings and Strategies}
\label{sec:memory_settings}
We have the following settings to simulate the strategies (including our proposed method) of dealing with stream data:
\begin{itemize}[noitemsep,nolistsep,leftmargin=*]
\item \textbf{All data}: Denoted as \textit{``All''}. In this setting, in each month, we use all available training data at that time point (e.g., in the $2^{nd}$ month we use training data from the $1^{st}$ and $2^{nd}$ months).
\item \textbf{Current batch only}: Denoted as \textit{``Current''}. In this setting, we exclusively use the \textit{current} month to train a new model for the current month.
\item \textbf{Aggregated model}: Denoted as \textit{``Finetune''}. In this setting, after training/finetuning a model using the data from the $i-1^{th}$ month, we iteratively finetune the model using the data from the $i^{th}$ month starting from the best model according to the F1-score of the development set.
\item \textbf{Proposed framework}: Denoted as \textit{``Proposed''}, our proposed method.
\end{itemize}

\subsection{Document Pre-Processing}
To preprocess documents, we use spaCy for segmentation, tokenization and PoS (Part-of-Speech) tagging.

Due to different tokenization tools, we do not use BERT for event trigger detection: when we use WordPiece tokenizer as BERT, the PoS tagger from spaCy does not work well; when we use spaCy tokenization results and input them in the pretrained BERT framework, the performance degrades significantly. Hence, in this paper, we use BERT to extract sentence-level embeddings only.

\subsection{Hyper parameters}
For the sentence-level embedding extractor, the parameters of BERT framework are pretrained\footnote{The model is available at \url{https://storage.googleapis.com/bert_models/2018_10_18/cased_L-12_H-768_A-12.zip}}. The \verb|CLS| output from BERT framework with regard to an input sentence is a $768$-dim vector\footnote{We did not use the $1024$-dim embedding due to limited RAM on GPU}, and the output dimension size of the sentence-level embedding is set to $256$ with a FC layer that project the $768$-dim vectors to $256$-dim vectors. We pretrain the parameters on this FC layer so that it works as a fixed sentence-level embedding extractor. We use the 360 example sentences in the ACE annotation guideline and use a Softmax layer with 33 (the number for event types) labels and train a sub-framework to predict the existence of an event type in the guideline sentences. After we remove the softmax layer, we consider the output from FC layer as a sentence-level embedding that captures the features indicating the existence of event types.

For the memory embedding extractor, the dimension size of the hidden state of the memory network is set to $128$.

For the core network, we use $200$-dim pre-trained \textbf{Word2Vec}~\cite{mikolov2013efficient} embeddings which are trained from Wikipedia article dump on January 1st, 2017 and tokenized with spaCy. $50$-dim PoS tag embeddings, and $20$-dim character embeddings (from a character-based Bi-LSTM network with the original input of $32$-dim character embedding and each direction has $10$-dim hidden state size). The Bi-LSTM that extracts the token's context embeddings is set to $128$-dim hidden state on both directions, with $256$-dim as the whole embedding size and a total of $384$ after being concatenated with memory embeddings. The FC layer before the CRF layer has an output dimension of $128$.

We have two optimizers in this framework (will be discussed in the following subsection). Both of them are Adam\cite{kingma2014adam}, with learning rate set to $0.001$.

\subsection{Train and Optimization}
In our work, we first pretrain the sentence-level embedding extractor from the FC layer as mentioned in Section 3.1. We use example sentences from ACE annotation guideline. We use 90\% as training sentences and 10\% as test sentences. To prevent over-fitting, we run the training process multiple times and select the model that converges (achieves highest test score) at the earliest epoch number.

The core network has another independent optimizer. The back-propagation from the core network will update both parameters on the Bi-LSTM-CRF framework~\cite{ma2016end} and the memory network; however, the sentence-level embedding extractor's parameters are fixed and not updated.

Conventional strategies (\textit{All}, \textit{Finetune}, \textit{Current}) do not train with memory network or sentence-level embedding extractor, they only go through the Bi-LSTM-CRF network (no memory embeddings are concatenated). Moreover, since there is no "previous" batch for the first batch, our proposed framework is trained from the second batch, and we inherit the parameters on Bi-LSTM-CRF network from conventional strategies and continue to train the parameters for both core and memory network as mentioned above. Moreover,at the second batch, the FC layer before CRF layer in our proposed framework is a newly initialized one with input dimension of $384$ and output of $128$ and later batches still follow this FC layer and its parameters, our empirical output did not see any negative impact with this approach.

\subsection{Results}
\paragraph{Performance} In Table~\ref{tab:performance_ace2005} we show the performance of the strategies on two noise levels. It is expected that the performance of the \textit{Current} strategy is the lowest among all strategies because the model does not have enough training data and easily over-fits; hence we observe a drop in the batch of 200305 even this batch has a higher quality of annotations. The \textit{Finetune} and \textit{All} strategies perform better, but the noise still influences the performance. The \textit{Finetune} strategy does not sufficiently alleviate the influence from wrong annotations in the early batches. Our method outperforms all other strategies because the memory embeddings successfully enhance the information with same event occurrence in the sentences in the past batches. Moreover, the improvement is more significant if the noise level is higher.

\paragraph{Complexity} We also demonstrate the average training time with different strategies in Table~\ref{tab:complexity}, we run the codes on 8 Intel CPU cores with $2.3$GHz and one Nvidia P100 GPU. The \textit{Finetune} and \textit{Batch} strategies spend a similar length of time and are faster than \textit{All} strategy because \textit{All} strategy takes data instances from both current and previous batches and this prolongs training time. The core network in our proposed framework processes the same amount of training instances as the \textit{Finetune} and \textit{Batch} strategies, with additional time for retrieving sentence-level embeddings and training the memorization module. Since the size of input to the memorization module is much smaller than all token embeddings from the previous batches, it takes less extra time than \textit{All} strategy. The cost for saved training time is small. Each sentence merely requires 6k bytes for storage if we use $768$-dim embedding and double floating-point.

\begin{table}[ht]
\centering
\small
\begin{tabular}{c|c}
\toprule
Time & Training Time (Seconds)\\
\midrule
\textit{All} & 18328\\
\textit{Current} & 12473\\
\textit{Finetune} & 12567\\
\textit{Proposed} & 14832\\
\bottomrule
\end{tabular}
\caption{Time consumption with different strategies.}
\label{tab:complexity}
\end{table}
\section{Related Work}
Most recent event extraction approaches, \textit{e.g.}, \cite{chen2018collective,liu2018jointly,hong2018self,nguyen2018graph,nguyen2018one,lu2018similar}, consider the training data as a whole batch, \cite{liu2018exploiting} also utilizes memory network for event extraction but it only considers the sentence from the same document and training instances come in a single batch. Our framework is the first one that tackles annotation arriving in batches.

Most approaches regard streaming data processing~\cite{ge2018fine,kuo2004event,Miranda2018Multilingual} focus on methods that \textit{output labels} on the data arriving in stream, while our framework \textit{consumes annotation} arriving in stream.

We use the term ``\textit{memory}'' to denote information recalled from past batches. We survey work such as \cite{weston2014memory,chu2018learning} where memory mechanism resembles attention while the memorization module in our framework follows the methods in \cite{sukhbaatar2015end,liu2018exploiting,ma2017long}, which are based on RNN or LSTMs.

Although we do not discuss in the previous sections, there are also other ways to work with the small batches of data sets, such as activate learning \cite{settles2009active,zhang2016name}. However, as we emphasized in Section~\ref{sec:introduction}, the annotators \textit{do not} iterate on the early batches of data set, hence we do not consider learning methods derived from those strategies as our baselines.

%The term ``\textit{Memory network}'' may refer to several different frameworks which share a property of locating information. Some approaches~\cite{chu2018learning} use attention mechanism to highlight information from source/reference portion of the data. Some frameworks~\cite{liu2018exploiting,ma2017long} establish memory networks based on LSTM network, which our work follows.
\section{Conclusion and Future Work}
In this paper, we propose a framework which recalls similar information in the past to handle a scenario where more annotation errors appear in early batches and annotators do not revisit those errors. This framework works well with the assumption that the annotation quality increases as the annotators will skill up in the later stage of annotation task. We use a memory embedding to emulate the progress and improvement of annotator's skills. In the future, we will focus on ``crowd-sourcing'' scenarios where the annotation quality is volatile across different batches due to unstable skills of ad-hoc annotators.

\section*{Acknowledgement}
This research is based upon work supported in part by U.S. DARPA AIDA Program No. FA8750-18-2-0014. The views and conclusions contained herein are those of the authors and should not be interpreted as necessarily representing the official policies, either expressed or implied, of DARPA, or the U.S. Government. The U.S. Government is authorized to reproduce and distribute reprints for governmental purposes notwithstanding any copyright annotation therein.

\bibliographystyle{unsrt}  
\bibliography{refs}

\end{document}